\documentclass[10pt,twocolumn,letterpaper]{article}

\usepackage[pagenumbers]{cvpr} 

\usepackage{graphicx}
\usepackage{amsmath}
\usepackage{amssymb}
\usepackage{booktabs}
\usepackage{multirow}
\usepackage{makecell}
\usepackage{algorithm}
\usepackage{algorithmic}
\usepackage{mathtools}
\usepackage{overpic}
\usepackage{url}
\usepackage{color}
\usepackage{colortbl}
\usepackage{xcolor}
\usepackage{pifont}
\usepackage{epstopdf}
\usepackage{svg}

\usepackage[pagebackref,breaklinks,colorlinks]{hyperref}

\newcommand{\nop}[1]{}

\definecolor{blue}{RGB}{164, 194, 244}

\usepackage[capitalize]{cleveref}
\crefname{section}{Sec.}{Secs.}
\Crefname{section}{Section}{Sections}
\Crefname{table}{Table}{Tables}
\crefname{table}{Tab.}{Tabs.}

\def\cvprPaperID{10135} 
\def\confName{CVPR}
\def\confYear{2023}

\begin{document}

\title{QCNeXt: A Next-Generation Framework For Joint Multi-Agent Trajectory Prediction}

\author{
Zikang Zhou$^{1, 2}$~~~
Zihao Wen$^{1, 2}$~~~
Jianping Wang$^{1, 2}$~~~
Yung-Hui Li$^3$~~~
Yu-Kai Huang$^4$~~~
\\
$^1$City University of Hong Kong~~~
$^2$City University of Hong Kong Shenzhen Research Institute~~~\\
$^3$Hon Hai Research Institute~~~
$^4$Carnegie Mellon University~~~
\\
{\tt\small zikanzhou2-c@my.cityu.edu.hk}\\
}
\maketitle

\begin{abstract}
Estimating the joint distribution of on-road agents' future trajectories is essential for autonomous driving. In this technical report, we propose a next-generation framework for joint multi-agent trajectory prediction called QCNeXt. First, we adopt the query-centric encoding paradigm for the task of joint multi-agent trajectory prediction. Powered by this encoding scheme, our scene encoder is equipped with permutation equivariance on the set elements, roto-translation invariance in the space dimension, and translation invariance in the time dimension. These invariance properties not only enable accurate multi-agent forecasting fundamentally but also empower the encoder with the capability of streaming processing. Second, we propose a multi-agent DETR-like decoder, which facilitates joint multi-agent trajectory prediction by modeling agents' interactions at future time steps. For the first time, we show that a joint prediction model can outperform marginal prediction models even on the marginal metrics, which opens up new research opportunities in trajectory prediction. Our approach ranks $1^{st}$ on the Argoverse 2 multi-agent motion forecasting benchmark, winning the championship of the Argoverse Challenge at the CVPR 2023 Workshop on Autonomous Driving.

\end{abstract}

\section{Introduction}
\label{sec:intro}

\begin{figure}[t]
  \centering
  \includegraphics[width=0.8\linewidth]{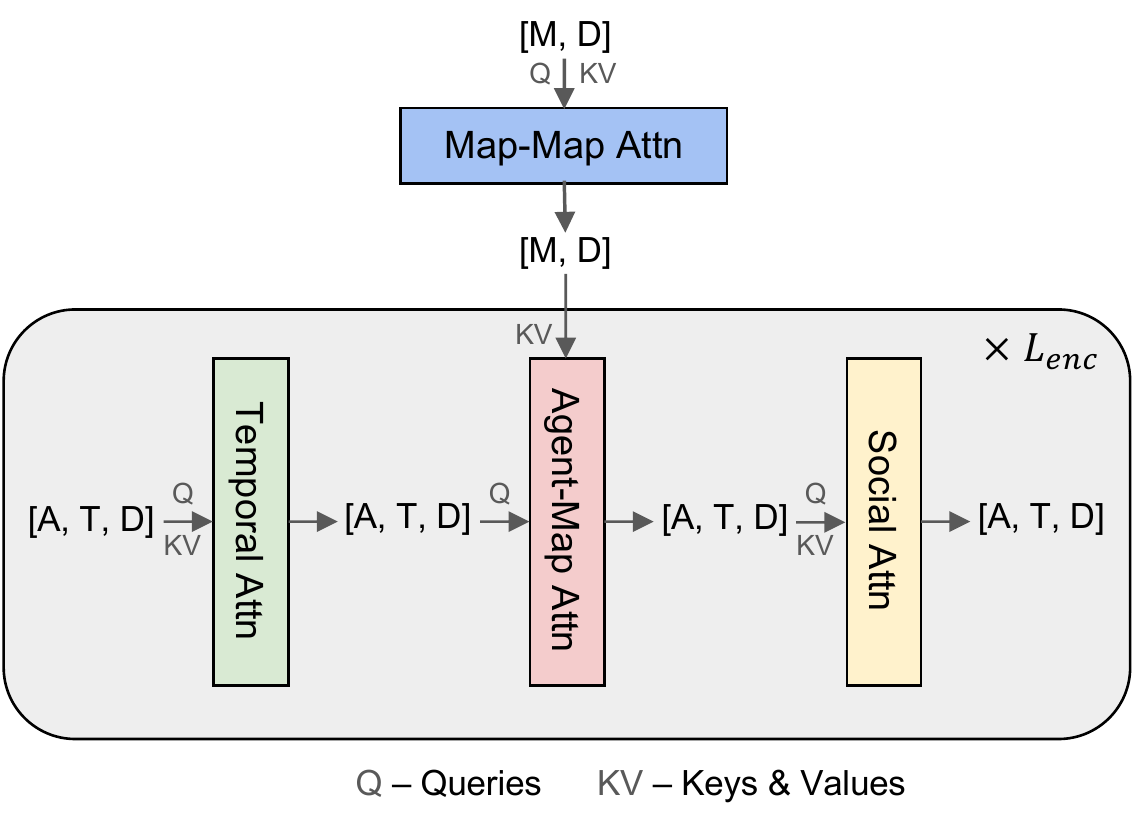}
      \caption{Overview of the scene encoder.}
  \label{fig:cubic_encoder}
\end{figure}

\begin{figure*}[t]
  \centering
  \includegraphics[width=1.0\linewidth]{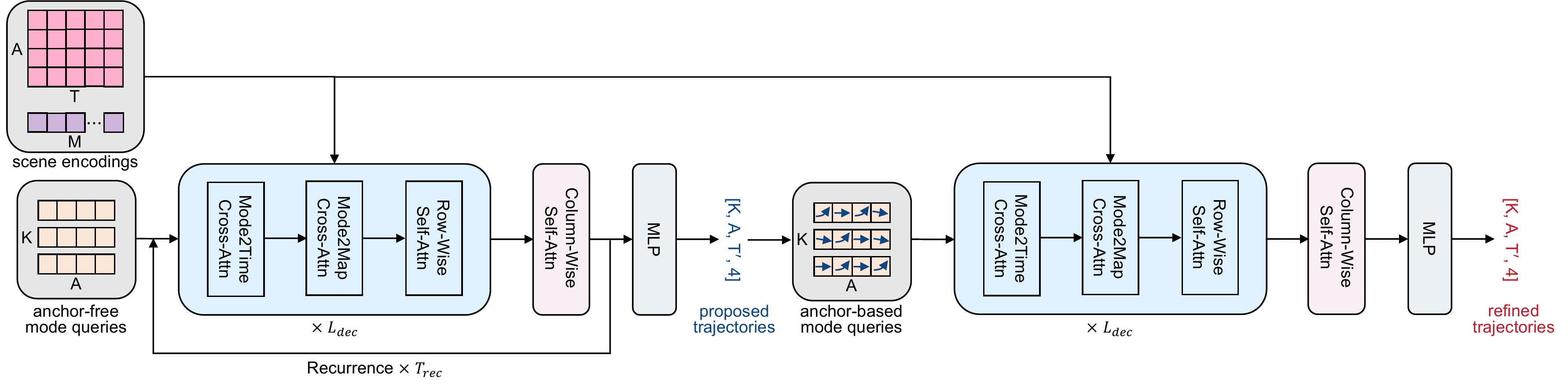}
      \caption{Overview of the decoding pipeline. An anchor-free module generates joint trajectory proposals \textit{recurrently} based on the encoded scene context. These proposals act as the anchors in the refinement module, where an anchor-based decoder refines the anchor trajectories.}
  \label{fig:joint_decoder}
\end{figure*}

Trajectory prediction is one of the toughest problems in autonomous driving. To enable safe autonomous driving, a trajectory prediction model must accurately forecast the future motions of one or multiple target agents (e.g., vehicles, pedestrians, cyclists, etc.) surrounding the autonomous vehicle. When it comes to multi-agent prediction, most previous works mainly focus on estimating the marginal distribution of target agents' future trajectories, i.e., assuming that the future movements of multiple agents are conditionally independent. This assumption may not be ideal for scene understanding and decision making in autonomous driving, as social interactions among agents take place not only at past time steps but also at future time steps. Some other works instead focus on joint multi-agent trajectory prediction by considering the future social interactions. However, none of these approaches can achieve the same level of performance as marginal prediction models on the marginal metrics. It was believed that the joint prediction task is much more difficult than the marginal prediction task~\cite{ngiam2021scene}.

In this technical report, we propose a next-generation modeling framework for joint multi-agent trajectory prediction, which can accurately estimate the joint future distribution of multiple target agents. We term this framework as QCNeXt since it is the next generation of QCNet~\cite{zhou2023query}, one of the most powerful marginal trajectory prediction models in the world. QCNeXt employs a Transformer-based encoder-decoder architecture as its predecessor. For the encoder, we inherit the symmetric design of HiVT~\cite{zhou2022hivt} and QCNet~\cite{zhou2023query}, which equips the model with permutation equivariance on the set elements, roto-translation invariance in the space dimension, and translation invariance in the time dimension. These invariance properties help the model achieve accurate multi-agent forecasting and enable streaming scene encoding. For the decoder, we extend QCNet's decoding pipeline to a joint prediction variant, which can capture agents' social interactions at future time steps explicitly. Furthermore, we introduce a scene scoring module to estimate the likelihood of all target agents' joint future trajectories. Experiments on the Argoverse 2 multi-agent motion forecasting benchmark~\cite{Argoverse2} demonstrate that QCNeXt can accurately forecast trajectories at the scene level. As a joint prediction model, QCNeXt can outperform QCNet even on the marginal metrics, which showcases the effectiveness of our solution.

\section{Approach}

\begin{table*}[t]
\footnotesize
\centering
\setlength{\tabcolsep}{1.4mm}
\begin{tabular}{lccccccc}
\toprule
Method & avgMinFDE${}_6$ $\downarrow$ & avgMinFDE${}_1$ $\downarrow$ & actorMR${}_6$ $\downarrow$ & avgMinADE${}_6$ $\downarrow$ & avgMinADE${}_1$ $\downarrow$ & \textbf{avgBrierMinFDE${}_6$ $\downarrow$} & actorCR${}_6$ $\downarrow$ \\
\midrule
testsss & 1.81 & 4.06 & 0.26 & 0.77 & 1.54 & 2.49 & 0.02 \\
FFINet & 1.77 & 3.18 & 0.24 & 0.77 & 1.24 & 2.44 & 0.02 \\
HistoryInfo & 1.86 & 3.62 & 0.25 & 0.84 & 1.44 & 2.40 & 0.02 \\
Forecast-MAE & 1.55 & 3.33 & 0.19 & 0.69 & 1.30 & 2.24 & 0.01 \\
\hline
QCNeXt (w/o ensemble) & 1.13 & 2.55 & 0.14 & 0.54 & 1.03 & 1.79 & 0.01 \\
QCNeXt (w/ ensemble) & \textbf{1.02} & \textbf{2.29} & \textbf{0.13} & \textbf{0.50} & \textbf{0.94} & \textbf{1.65} & 0.01 \\
\bottomrule
\end{tabular}
\caption{Quantitative results on the Argoverse 2 multi-agent motion forecasting leaderboard.}
\label{tab:av2_joint_test}
\end{table*}

\begin{table}[t]
\footnotesize
\centering
\setlength{\tabcolsep}{2.2mm}
\begin{tabular}{lcccc}
\toprule
Method & b-minFDE${}_6$ $\downarrow$ & minADE${}_6$ $\downarrow$ & minFDE${}_6$ $\downarrow$ & MR${}_6$ $\downarrow$ \\
\midrule
QCNet & 1.90 & 0.64 & 1.28 & 0.16 \\
QCNeXt & 1.90 & 0.64 & \textbf{1.24} & \textbf{0.15} \\
\bottomrule
\end{tabular}
\caption{Quantitative results on the Argoverse 2 validation set. We evaluate the scenes where only one agent is required to be predicted. All the metrics are for marginal trajectory prediction.}
\label{tab:av2_joint_val}
\end{table}

This section describes the framework of QCNeXt in detail. First, we briefly introduce the scene encoder used by our model. Second, we elaborate on our proposed decoder, which is specially designed for the task of joint multi-agent trajectory prediction. Next, we explain how scene-level scoring is conducted in our framework. Finally, we illustrate the training objectives used for optimizing a joint prediction model.

\subsection{Query-Centric Scene Encoder}

Our scene encoder is the same as the one used in QCNet~\cite{zhou2023query}, which is a factorized attention-based Transformer that captures temporal dependencies, agent-map interactions, and social interactions. The overall architecture of the encoder is shown in \cref{fig:cubic_encoder}. We adopt the query-centric paradigm in QCNet to encode the scene elements. The philosophy behind this encoding paradigm is relative spacetime, which guides us to equip models with roto-translation invariance in the space dimension and translation invariance in the time dimension. In this paradigm, a local spacetime coordinate system is built for each scene element, including lanes, crosswalks, vehicles, pedestrians, etc. These scene elements are then encoded in their local coordinate systems to produce invariant representations, and the relationships between scene elements are captured by Transformers with the help of relative spatial-temporal positional embeddings. Specifically, the key/value elements in attention layers are concatenated with the spatial-temporal positional embeddings relative to the query elements before performing QKV attention. Following the map-map attention as well as a series of temporal attention, agent-map attention, and social attention, the scene encoder produces map encodings of shape $[M, D]$ and agent encodings of shape $[A, T, D]$, where $M$, $A$, $T$, $D$ are the numbers of map polygons, modeled agents, historical time steps, and hidden units, respectively. These encodings will later serve as the scene context in the decoder. For more details about the scene encoder, please kindly refer to the QCNet paper~\cite{zhou2023query}.

\subsection{Multi-Agent DETR Decoder}

Our decoding pipeline follows the design choices of QCNet's decoder, where a recurrent, anchor-free trajectory proposal module generates adaptive trajectory anchors in a data-driven manner, followed by an anchor-based trajectory refinement module that predicts the offset to the trajectory anchors. However, the original decoder of QCNet does not consider the social interactions among agents at future time steps since it only aggregates neighboring agents' encodings at the current time step. As a result, the QCNet decoder is only suitable for marginal trajectory prediction. To address this issue, we propose a new DETR-like decoder that can capture the future social interactions. The detailed architecture of our decoder is shown in \cref{fig:joint_decoder}.

\noindent \textbf{Anchor-Free Trajectory Proposal.} We randomly initialize $K$ embeddings with the size of $D$ before training. Each of these embeddings is then repeated $A$ times to form a tensor of shape $[K, A, D]$, where each row serves as the initial seed of one of the $K$ joint futures. For each row of this tensor, the $A$ embeddings are first updated with a Mode2Time cross-attention module, which makes each embedding take charge of the prediction for one of the agents in the scene. Then, a Mode2Map cross-attention module updates the embeddings with neighboring map information. Next, we apply row-wise self-attention on the embedding tensor, which aims to model the social interactions among agents within each joint scene. These three modules are stacked $L_{\text{dec}}$ times interleavely, followed by a column-wise self-attention module that enables the $K$ joint scenes to communicate with each other. An MLP is then used to decode $2$ seconds' trajectories from the updated embedding tensor. To predict the next $2$ seconds' trajectories, we let the updated embedding tensor become the input of the Mode2Time cross-attention module again and repeat the aforementioned process. This computing process is conducted recurrently until the $6$-second trajectories are completed.

\noindent \textbf{Anchor-Based Trajectory Refinement.} The trajectories predicted by the proposal module serve as the anchors for the refinement module. Compared with the proposal module, the refinement module's initial embedding tensor is not learnable but is derived from the trajectories output by the proposal module. The remaining architecture is similar to that of the proposal module, except that the MLP predictor outputs the offset to the trajectory anchors in a single shot without using any recurrent mechanism.

\subsection{Scene Scoring Module}

In contrast to QCNet's decoder which produces agent-level trajectory scores via an MLP that applies on the post-refinement mode embeddings, our decoder needs to generate scene-level confidence scores to adapt to the joint trajectory prediction task. The scene scoring module takes as input the post-refinement mode embeddings of all target agents in the scene. To generate one confidence score per joint prediction, we need some scene-level pooling operators to summarize all target agents' mode embeddings into one scene embedding and decode the confidence score from it via an MLP. Typical pooling methods include average pooling, max pooling, attentive pooling, etc. We empirically choose attentive pooling as we notice that some target agents have uninteresting behavior (e.g., keeping static) and should not contribute too much to the calculation of the scene score.

\subsection{Training Objectives}

We parameterize the joint future trajectory distribution of all target agents as a mixture of Laplace distributions, which has the form of
\begin{equation}
\sum_{k=1}^K \pi_{k} \prod_{i=1}^{A^{\prime}} \prod_{t=1}^{T^{\prime}} f\left(\mathbf{p}_i^{t, x} \mid \boldsymbol{\mu}_{i, k}^{t, x}, \mathbf{b}_{i, k}^{t, x}\right) f\left(\mathbf{p}_i^{t, y} \mid \boldsymbol{\mu}_{i, k}^{t, y}, \mathbf{b}_{i, k}^{t, y}\right).
\label{eq:dist}
\end{equation}
In this formulation, $f(\cdot \mid \cdot)$ denotes the probability density function of Laplace distribution, and $\{\pi_{k}\}_{k=1}^{K}$ are the mixing coefficients of the multi-agent, multi-step Laplace components. The dimension of each Laplace component is $A^{\prime} \times T^{\prime} \times 2$, where $A^{\prime}$ is the number of target agents to be predicted ($A^{\prime} \leq A$) and $T^{\prime}$ is the number of future time steps ($T^{\prime} = 60$ for the Argoverse 2 dataset). For the $i$-th agent in the $k$-th mixture component, its mean position at time step $t$ is parameterized by the location $\boldsymbol{\mu}_{i, k}^t = (\boldsymbol{\mu}_{i, k}^{t, x}, \boldsymbol{\mu}_{i, k}^{t, y})$, and the corresponding uncertainty score is parameterized by the scale $\mathbf{b}_{i, k}^t = (\mathbf{b}_{i, k}^{t, x}, \mathbf{b}_{i, k}^{t, y})$. We use a classification loss $\mathcal{L}_{\text{cls}}$ to optimize the mixing coefficients, which minimizes the negative log-likelihood of \cref{eq:dist}. Moreover, we employ the scene-level winner-take-all strategy~\cite{ngiam2021scene} to minimize the negative log-likelihood of the winner mode's Laplace distribution. The winner mode is defined as the best scene-level trajectory proposal and its refinement, where the best scene-level trajectory proposal has the minimum displacement error with respect to the ground truth. The final loss function combines the trajectory proposal loss $\mathcal{L}_{\text{propose}}$, the trajectory refinement loss $\mathcal{L}_{\text{refine}}$, and the classification loss $\mathcal{L}_{\text{cls}}$:
\begin{equation}
\mathcal{L}=\mathcal{L}_{\text{propose}}+\mathcal{L}_{\text{refine}}+ \mathcal{L}_{\text{cls}}\,.
\label{eq:loss}
\end{equation}

\section{Experiments}

\subsection{Implementation Details}

The hidden feature dimension is $128$. All layers for information fusion have the same architecture, which is similar to the gated variant of the attention mechanism used in HiVT~\cite{zhou2022hivt}. All multi-head attention layers use $8$ heads. For training stability, we detach the gradients of the proposed trajectory anchors before the anchors are used in the refinement module. We use the AdamW optimizer~\cite{loshchilov2017decoupled} for training. The training process lasts $50$ epochs with a batch size of $32$. Both the dropout rate and the weight decay rate are set to $0.1$. The learning rate is decayed from $5 \times 10^{-4}$ to $0$ by the cosine annealing scheduler~\cite{loshchilov2016sgdr}. We use $3$ recurrent steps in the trajectory proposal module and $2$ blocks of multi-context attention in both the encoder and the decoder.

\subsection{Ensembling}

We use different random seeds to train $8$ models, which produces $48$ scene-level predictions in total. For each scenario, the $48$ scene-level predictions are used for ensembling based on the weighted k-means algorithm. Specifically, the joint endpoints of all target agents in the scene are taken as the input of the weighted k-means algorithm, and the scene-level scores are used as sample weights. After cluster assignment, the joint trajectories within each cluster are averaged. This can be viewed as a simple extension of the commonly used ensembling strategies for marginal trajectory prediction.

\subsection{Quantitative Results}

The performance of QCNeXt on the Argoverse 2 multi-agent motion forecasting benchmark is shown in \cref{tab:av2_joint_test}, where we can see that our ensembling strategy can significantly boost the model's performance. But even without using ensembling, our approach has already outperformed all methods on all metrics by a significant margin, which demonstrates the superiority of our modeling framework.

We found that about 20\% of the scenarios in the Argoverse 2 validation/test set evaluate only one agent's prediction result. In this case, the formulations of joint trajectory distribution and marginal trajectory distribution become equivalent, so we are curious about the performance comparison between joint prediction models and marginal prediction models in these scenarios. Previously, the literature believed that joint prediction models cannot achieve the same level of performance as marginal prediction models on the marginal metrics, given that the joint prediction task must account for the consistency among agents' future trajectories and is thus a more challenging task~\cite{ngiam2021scene}. However, we surprisingly found that this conclusion does not hold for our approach. As shown in \cref{tab:av2_joint_val}, QCNeXt can perform better than QCNet~\cite{zhou2023query}, the most powerful marginal prediction model on Argoverse 2, on marginal metrics like minFDE${}_6$ and MR${}_6$. We attribute this amazing result to the effectiveness of our modeling framework.

\section{Conclusion}

We propose QCNeXt, an elegant and performant framework for joint multi-agent trajectory prediction. Inherited from the symmetric designs of HiVT~\cite{zhou2022hivt} and QCNet~\cite{zhou2023query}, our modeling framework easily achieves fast and accurate multi-agent motion forecasting. By explicitly modeling the future relationships among agents with a multi-agent DETR-like decoder, the joint future motions of multiple agents are predicted more precisely. Experiments show that our method achieves state-of-the-art performance on the Argoverse 2 multi-agent motion forecasting benchmark.

\section*{Acknowledgement}

This work was partially supported by Hong Kong Research Grant Council under GRF 11200220, Science and Technology Innovation Committee Foundation of Shenzhen under Grant No. JCYJ20200109143223052.


{\small
\bibliographystyle{ieee_fullname}
\bibliography{egbib}

\begin{thebibliography}{1}\itemsep=-1pt

\bibitem{loshchilov2016sgdr}
Ilya Loshchilov and Frank Hutter.
\newblock Sgdr: Stochastic gradient descent with warm restarts.
\newblock In {\em Proceedings of the International Conference on Learning
  Representations (ICLR)}, 2017.

\bibitem{loshchilov2017decoupled}
Ilya Loshchilov and Frank Hutter.
\newblock Decoupled weight decay regularization.
\newblock In {\em Proceedings of the International Conference on Learning
  Representations (ICLR)}, 2019.

\bibitem{ngiam2021scene}
Jiquan Ngiam, Benjamin Caine, Vijay Vasudevan, Zhengdong Zhang, Hao-Tien~Lewis
  Chiang, Jeffrey Ling, Rebecca Roelofs, Alex Bewley, Chenxi Liu, Ashish
  Venugopal, et~al.
\newblock Scene transformer: A unified architecture for predicting multiple
  agent trajectories.
\newblock In {\em Proceedings of the International Conference on Learning
  Representations (ICLR)}, 2022.

\bibitem{Argoverse2}
Benjamin Wilson, William Qi, Tanmay Agarwal, John Lambert, Jagjeet Singh,
  Siddhesh Khandelwal, Bowen Pan, Ratnesh Kumar, Andrew Hartnett,
  Jhony~Kaesemodel Pontes, Deva Ramanan, Peter Carr, and James Hays.
\newblock Argoverse 2: Next generation datasets for self-driving perception and
  forecasting.
\newblock In {\em Proceedings of the Neural Information Processing Systems
  Track on Datasets and Benchmarks (NeurIPS Datasets and Benchmarks)}, 2021.

\bibitem{zhou2023query}
Zikang Zhou, Jianping Wang, Yung-Hui Li, and Yu-Kai Huang.
\newblock Query-centric trajectory prediction.
\newblock In {\em Proceedings of the IEEE/CVF Conference on Computer Vision and
  Pattern Recognition (CVPR)}, 2023.

\bibitem{zhou2022hivt}
Zikang Zhou, Luyao Ye, Jianping Wang, Kui Wu, and Kejie Lu.
\newblock Hivt: Hierarchical vector transformer for multi-agent motion
  prediction.
\newblock In {\em Proceedings of the IEEE/CVF Conference on Computer Vision and
  Pattern Recognition (CVPR)}, 2022.

\end{thebibliography}
}

\end{document}